%% file: emnlp2022.tex
\pdfoutput=1
\documentclass[11pt]{article}

\usepackage[]{EMNLP2022}

\usepackage{times}
\usepackage{latexsym}

\usepackage[T1]{fontenc}
\usepackage[utf8]{inputenc}

\usepackage{graphicx}
\usepackage{amsmath}
\usepackage{breqn}
\usepackage{makecell}
\usepackage{tabularx}
\usepackage{booktabs}
\usepackage{multirow}
\usepackage[colorinlistoftodos]{todonotes}
\usepackage{textcomp}
\usepackage{bm}

\usepackage{hyperref}
\usepackage{amsmath}
\usepackage{multirow}

\usepackage{microtype}

\usepackage{inconsolata}

\let\oldfootnote\footnote
\def\footnote{\ifhmode\unskip\fi\oldfootnote}

\title{Improving Factual Consistency in Summarization \\ with Compression-Based Post-Editing}
\author{\quad Alexander R. Fabbri \quad Prafulla Kumar Choubey \quad Jesse Vig \\
 {\bf Chien-Sheng Wu \quad \stepcounter{footnote}Caiming Xiong} \\
Salesforce AI Research \\
\texttt{\{afabbri, pchoubey, jvig, wu.jason, cxiong\}@salesforce.com} \\
}

\begin{document}
\maketitle

\input{0-abstract}
\input{1-introduction}
\input{2-methodology}
\input{3-results}
\input{4-conclusion}
% \clearpage
\input{5-ethical-considerations}
% \clearpage
\bibliography{anthology,custom}
\bibliographystyle{acl_natbib}
\appendix
\input{6-appendix}
\end{document}

%% file: 0-abstract.tex
\begin{abstract}
State-of-the-art summarization models still struggle to be factually consistent with the input text.
A model-agnostic way to address this problem is post-editing the generated summaries.
However, existing approaches typically fail to remove entity errors if a suitable input entity replacement is not available or may insert erroneous content. 
In our work, we focus on removing extrinsic entity errors, or entities not in the source, to improve consistency while retaining the summary's essential information and form. 
We propose to use sentence-compression data to train the post-editing model to take a summary with extrinsic entity errors marked with special tokens and output a compressed, well-formed summary with those errors removed. 
We show that this model improves factual consistency while maintaining ROUGE, improving entity precision by up to 30\% on XSum, and that this model can be applied on top of another post-editor, improving entity precision by up to a total of 38\%.
% 
% (80.02 - 61.59)/61.59 = 0.2992
% (85.07 - 61.59)/61.59 = 0.3812
% 
We perform an extensive comparison of post-editing approaches that demonstrate trade-offs between factual consistency, informativeness, and grammaticality, and we analyze settings where post-editors show the largest improvements. 
\end{abstract}

%% file: 1-introduction.tex
\section{Introduction}\label{sec:introduction}
% 
\input{tables/summcomp_figure}
Text summarization aims to compress a long document(s) into a short and fluent form that preserves salient information. 
State-of-the-art models, however, are often not factually consistent with the input they are conditioned on~\cite{maynez-etal-2020-faithfulness,TACL2563}.
%
% 
% Recent work has focused on improving metrics for factual consistency \cite{maynez-etal-2020-faithfulness,10.1162/tacl_a_00453,fabbri-etal-2022-qafacteval} and modeling approaches \cite{nan-etal-2021-entity,kang-hashimoto-2020-improved,aralikatte2021focus}. 
While recent modeling techniques have been proposed to improve factual consistency \cite{nan-etal-2021-entity,kang-hashimoto-2020-improved,aralikatte2021focus}, a model-agnostic approach is to post-edit the summaries.
\par
Prior work in post-editing for factual consistency has focused on swapping inconsistent entities with those in the input \cite{dong-etal-2020-multi,https://doi.org/10.48550/arxiv.2204.08263}, including by reranking the entity-replaced summaries \cite{chen-etal-2021-improving}, autoregressive approaches that learn to rewrite and remove perturbations from the input \cite{cao-etal-2020-factual,zhu-etal-2021-enhancing,adams2022redress}, or deletion-based editing of references \cite{wan2022factpegasus}. 
We provide example outputs of these approaches that demonstrate potential downsides in Table \ref{tab:example_intro}.
\input{tables/example_intro}
For summaries in datasets such as XSum \cite{narayan-etal-2018-dont} that contain extrinsic entity errors in up to 70\% of reference and generated summaries \cite{maynez-etal-2020-faithfulness}, \textit{a suitable entity replacement is often not available} from the source. 
% 
% In such cases, the entity swap-based approaches leave the summary unchanged.
% 
% Other approaches may modify the summary meaning by introducing potentially erroneous information in the case of rewriting-based approaches, or leave the summary ungrammatical in the case of deletion approaches. 
% \af{I removed the above two sentences since the content is mostly repeated in the Table}
% 
% 
\par
To address such errors, we propose a post-editing model that performs \textit{controlled compression}: given input text and a set of (factually incorrect) entities, it generates a compressed output with the specified entities removed, as demonstrated at the bottom of Table \ref{tab:example_intro}.
To synthesize a training set for this model, we use existing sentence compression data to first train a perturber model that maps a compressed sentence and a list of entities to an uncompressed sentence containing those entities.
% %
We then apply the perturber model to insert entities on a subset of the target dataset whose summaries contain only named entities also found in the input.
%  not containing extrinsic errors, as determined by named entity overlap
% % 
Our post-editor is then trained in the reverse direction, conditioning upon the input article and the perturbed reference summary, with entities to remove marked with special tokens, to produce the compressed summary, as illustrated in Figure \ref{tab:summ_comp}.
\par
Our contributions are the following: 
1) We propose a compression-based method for summary post-editing that removes extrinsic entity errors, improving entity precision with respect to the input by up to 25\% along with improvements in other factual consistency metrics while retaining informativeness according to automatic analysis. 
2) We show that this method can be applied on top of a rewriting-based post-editor, improving entity precision by up to 38\% overall with a small decrease in ROUGE score.  
3) We perform an extensive comparison of prior post-editing methods across two datasets and six summarization models to better understand the trade-offs between factual consistency, informativeness, and grammaticality. 
Models are made publicly available: \url{https://github.com/salesforce/CompEdit}.

%% file: tables/summcomp_figure.tex
%%%% Overview both, two columns
\begin{figure*}[t]
    \centering
    \includegraphics[width=.80\linewidth]{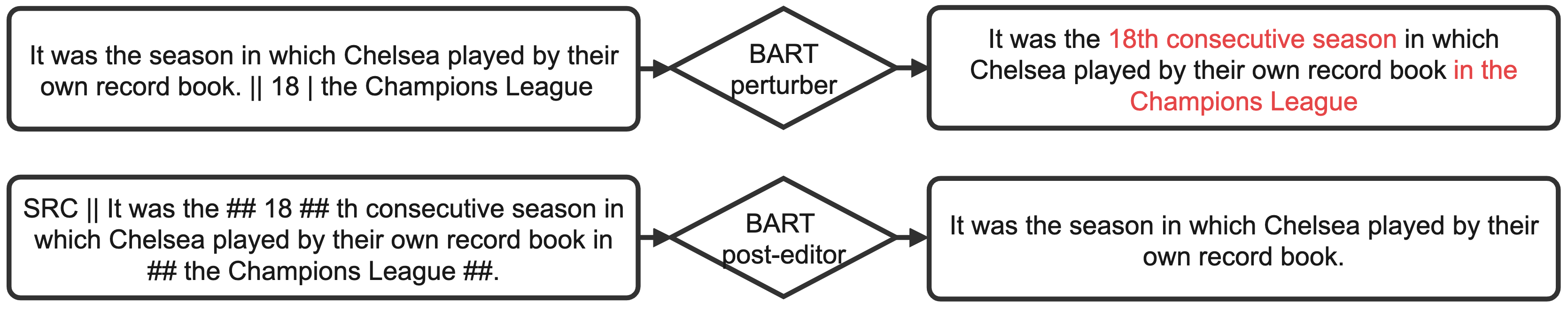}
    \caption{Overview of our compression-based approach. A perturber model is trained to produce an uncompressed sentence with inserted entities, conditioned upon the compressed sentence and an appended list of entities. The perturber is applied to a dataset's reference summaries to create data to train a post-editor that takes the document (SRC) and a summary with entities to remove surrounded by special tokens, and removes the erroneous words.}
    \label{tab:summ_comp}
\end{figure*}
% 
% 
%%%% Overview post-editor only, two columns
% \begin{figure*}
%     \centering
%     \includegraphics[width=\linewidth]{figures/overview_editor.png}
%     \caption{Overview of our compression-based approach. Our post editor takes the source document (SRC) and a summary with entities to remove surrounded by special tokens and learns to remove the erroneous words.}
%     \label{ftab:summ_comp}
% \end{figure*}
% 

%%%% Overview both, single column
% \begin{figure}
%     \centering
%     \includegraphics[width=0.5\textwidth]{figures/overview.png}
%     \caption{Overview of our compression-based approach. A perturber model is trained to produce an uncompressed sentence conditioned upon the compressed sentence and an appended list of entities. Our post editor takes the source document (SRC) and a summary with entities to remove surrounded by special tokens and learns to remove the erroneous words.}
%     \label{ftab:summ_comp}
% \end{figure}

%% file: tables/example_intro.tex
\begin{table}
\small
\centering
\resizebox{\columnwidth}{!}{\begin{tabular}{|p{0.20\linewidth}|p{0.80\linewidth}|}
\hline 
\multicolumn{2}{|p{\linewidth}|} {\textbf{Source}: The oil price collapse sent global markets reeling throughout 2015. ... Brent crude oil  was up 3\% at \$37.60 per barrel for the day but down 35\% over the year. } \\ \hline
System & \multicolumn{1}{c|}{Summary} \\ \hline 
Original & Wall Street markets closed lower on the \textcolor{red}{last trading day of 2015} as oil prices languished at \textcolor{red}{\$28} a barrel for much of the year. \\  \hline
Entity Swap &  \multicolumn{1}{|c|}{No Change} \\ \hline
Rewrite & Wall Street markets closed lower on the \textcolor{red}{last trading day of 2015} as oil prices languished at \$37.60 per barrel ... but remain down 35\% over the year. \\ \hline
Delete & Wall Street markets closed lower \textcolor{orange}{on 2015} as oil prices languished \textcolor{orange}{at a} for much of the year.  \\  \hline
Compress (ours) &   Wall Street markets closed lower as oil prices languished for much of the year. \\ \hline
\end{tabular}}
\caption{Example of post-editing approaches. Entity swap \cite{chen-etal-2021-improving} does not modify the errors (red) if a suitable entity replacement is not found, revision-based editing \cite{adams2022redress} performs more extensive changes, deletion \cite{wan2022factpegasus} removes errors but affects grammaticality (orange), while ours returns a well-formed summary through compression.}
\label{tab:example_intro}
\end{table}

%% file: 2-methodology.tex
\section{Methodology}\label{sec:methodology}
\input{tables/results_main}

\subsection{Prior Post-Editors}
\par
\noindent \textbf{SpanFact} We implement a variation of the autoregressive model from \citet{dong-etal-2020-multi}. 
We train a BART-large \cite{lewis-etal-2020-bart} model to take the source document and the summary with entity slots masked and fill in the masked slots. 
% produce the reference summary conditioned on the concatenation of the source and the reference summary with all entity slots masked. 
% 
We also train the model on the data subset whose summaries contain only named entities found in the input, which we call \textbf{SpanFact-c}.
\par
\noindent \textbf{CCGS} We apply the model from \citet{chen-etal-2021-improving} that generates candidate summaries by enumerating all ways to replace summary entities with similar-typed input entities and training BART with a classification layer to re-rank these summaries.
\par
\noindent \textbf{ReDRESS} We apply the approach from \citet{adams2022redress} for revising clinical reference summaries to news summarization. 
The approach consists of two stages: 1) a perturber learns to corrupt a summary by using entity swaps between the input and a retrieved set of entities, span deletion, and shuffling as training data. 2) The perturber is then applied to the reference summaries to create training samples for a reviser that learns to remove errors through contrastive learning. 
In contrast, our approach focuses on more controlled perturbations and revisions to remove and compress rather than rewrite, which may insert errors.
\par
\noindent \textbf{FactPegasus} We apply the deletion-based corrector component of \citet{wan2022factpegasus}.
This method removes extrinsic entity error tokens and surrounding words based on manually-defined rules over the dependency parse of the summary, which may introduce grammatical errors. 
\subsection{Compression-Based Post-Editor}
We train a post-editor in two steps on sentence compression data from \citet{filippova-altun-2013-overcoming}, where the uncompressed sentence can be viewed as the summary containing information not present in the compressed version.
Example input and output for these two steps are shown in Figure \ref{tab:summ_comp}.
We first train a BART-large \textbf{perturber} model conditioned upon a compressed sentence and entities present in the longer sentence but not present in the compressed one to produce a longer sentence. In other words, we maximize the following probability: $P(\text{Uncompressed}\; |\; \text{Compressed},\; \text{Ents})$.
Then, for a given data point in the summarization dataset, we select each of one, two, and three entities from the source not present in the reference and condition upon those entities and the reference summary to produce a longer, perturbed version containing those entities.
Varying the number of entities inserted mirrors differing levels of editing required.
Note that while the entities inserted are not extrinsic errors since they appear in the source, they are inserted out-of-context since the perturber does not condition upon the source document.
We then train our BART-large model \textbf{post-editor} model, which we call \textbf{CompEdit}, to produce the original reference summaries conditioned upon the source document and these perturbed summaries, with special tokens surrounding the entities that should be removed.
In this setup, the post-editor learns to remove related, out-of-context entities during training. 
We only train on references for which all named entities were found in the source and thus maximize the following: P(Gold summary | source, special-token perturbed summary).
While a post-editor could be trained directly from sentence compression data or on summarization data without access to the source, such models resulted in a degradation of summary salience.
During inference, we surround extrinsic entity errors, as determined by named entity overlap with the input, with special tokens to signal the model to remove them.

%% file: tables/results_main.tex
\begin{table*}[t!]
\small
\centering
\setlength{\tabcolsep}{2pt}
\resizebox{\linewidth}{!}{\begin{tabular}{| l | c c  c  c | c c c c c c | c | c |}
\hline
Model & E-P$_{src}$ & BS-P$_{src}$ & D$_{arc}$ & QAFE  & E-R$_{ref}$ & BS-F1$_{ref}$ & R1 & R2 & RL & R1-c & CoLA & Edit\% \\ \hline
% \multicolumn{13}{|c|}{XSUM} \\ \hline
BART & 61.59 & 41.69 & 82.80 & 1.960 & 55.02 & 48.88 & 45.20 & 21.95 & 36.95 & 43.59 & 98.83 & - \\
BART-c & 80.42 & 44.39 & 89.53 & 2.161 &  41.47 & 44.83 & 41.10 & 17.63 & 33.10 & 41.32 & 98.81 & - \\ \hline
SpanFact \tiny \cite{dong-etal-2020-multi} & 61.96 & 41.70 & 82.90 & 1.947  & \textbf{54.51} & \textbf{48.77} & \textbf{45.10}  & \textbf{21.88} & \textbf{36.88} & 43.54 & \textbf{98.77} & 40.53 \\
SpanFact-c & 74.26 & 42.82 & 85.85 & 1.978 & 45.20 & 47.32 & 43.44 & 19.80 & 35.42 & 43.45 & 98.44 & 60.27\\
CCGS \tiny \cite{chen-etal-2021-improving} & 64.80 & 41.93 & 83.44 & 1.938  & \textbf{53.22} & \textbf{48.56} & \textbf{44.89} & \textbf{21.61} & \textbf{36.71} & 43.57 & 98.54 & 15.07 \\
ReDRESS \tiny \cite{adams2022redress} & 75.42 & \textbf{44.90} & 88.54 & \textbf{2.168} & 47.37 & 46.98 & 43.30 & 20.12 & 35.29 & 42.81 & \textbf{98.79} & 61.82 \\
FactPegasus \tiny \cite{wan2022factpegasus} & \textbf{98.71} & 42.82 & \textbf{90.82} & 2.082 & 35.76 & 44.39 & 41.98 & 18.30 & 34.45 & \textbf{43.98} & 93.25 & \textbf{67.05} \\
CompEdit (ours) & 80.02 & 43.03 & 88.35 & 2.124  & 44.49 & 47.00 & 43.45 & 20.29 & 35.63 & \textbf{43.61} & 98.61 & 54.31 \\
ReDRESS + CompEdit & \textbf{85.07} & \textbf{45.39} & \textbf{90.61} & \textbf{2.224} & 42.20 & 45.97 & 42.32 & 19.19 & 34.53 & 42.67 & 98.66 & \textbf{73.36} \\ \hline
% \multicolumn{13}{|c|}{CNN/DM} \\
% \hline
% BART & 98.50 & 71.10 & 98.57 & 4.550 & 58.93 & 43.55 & 44.05 & 21.12 & 41.02 & 44.11 & 95.68 & -  \\
% BART-c & 97.60 & 68.59 & 97.88 & 4.405 & 58.00 & 43.70 & 44.09 & 20.89 & 41.11 & 44.14 & 95.11 & -  \\ \hline
% SpanFact \tiny \cite{dong-etal-2020-multi} & 98.46 & 71.06 & 98.49 & 4.545 & 58.32 & 43.45 & 43.97 & 21.02 & 40.95 & 44.03 & \textbf{95.60} & 28.29 \\
% SpanFact-c & 98.83 & 71.07 & 98.53 & 4.544 & 58.19 & 43.41 & 43.95 & 20.99 & 40.93 & 44.01 & 95.50 & 32.36 \\
% CCGS \tiny \cite{chen-etal-2021-improving} & 98.18 & 71.04 & 98.49 & 4.466  & 58.00 & 43.48 & \textbf{43.98} & 21.03 & \textbf{40.96} & 44.04 & 95.48 & 14.93 \\
% ReDRESS \tiny \cite{adams2022redress} & 99.03 & \textbf{71.32} & 98.66 & \textbf{4.557}  & \textbf{58.81} & \textbf{43.53} & \textbf{43.98} & \textbf{21.08} & \textbf{40.96} & 44.05 & \textbf{95.66} & \textbf{30.13} \\
% FactPegasus \tiny \cite{wan2022factpegasus} & \textbf{99.94} & 71.10 & 98.68 & 4.543  & 58.40 & 43.44 & \textbf{43.98} & 21.03 & \textbf{40.96} & \textbf{44.12} & 95.21 & 13.44 \\
% CompEdit (ours) & \textbf{99.10} & 71.13 & \textbf{98.75} & 4.554 & \textbf{58.61} & \textbf{43.50} & \textbf{44.02} & \textbf{21.08} & \textbf{41.01} & \textbf{44.11} & 95.52 & 27.93 \\
% ReDRESS + CompEdit & 98.95 & \textbf{71.24} & \textbf{98.74} & \textbf{4.555} & 58.49 & 43.44 & 43.96 & 21.04 & 40.95 & 44.04 & 95.42  & \textbf{44.53} \\ \hline
\end{tabular}}
\caption{Baseline and post-editing automatic results for factual consistency, relevance, and grammaticality metrics on XSum. The top two scores from the post-editors in each column are highlighted. 
% \jw{if needs more space we can put CNN results to appendix}
} \label{table-main-result}
\end{table*}

%% file: 3-results.tex
\section{Experiments}\label{sec:results}
\subsection{Settings}
We evaluate the above approaches on the XSUM \citep{narayan-etal-2018-dont} and CNN/DM \citep{DBLP:journals/corr/HermannKGEKSB15} datasets.
We apply BART-large as the base summarization model, and post-editing models are applied to this base model's output.
We also report results for BART-large trained on a data subset in which all summary named entities are found in the source, which we call \textbf{BART-c}.
All SpanFact and CompEdit models are trained for 10 epochs with a batch size of 64, with the best checkpoint chosen according to the highest average of ROUGE-1/2/L \cite{lin-2004-rouge} on the validation set. 
Additional dataset and modeling details, including CNN/DM results, are found in the Appendix.
\par
To show the generalizability of our results to post-editing models other than BART, we test the best post-editors on UniLM \cite{unilm} and  BottomUp \cite{gehrmann-etal-2018-bottom} outputs (additional models in the Appendix).
\citet{zhu-etal-2021-enhancing} released the output of their post-editor \textbf{FASumFC} on these outputs but not their trained model. 
This post-editor is a seq2seq denoising model trained with entity swaps for inserting inconsistencies and backtranslation to create paraphrases.
\input{tables/fasum_results_main}

\subsection{Automatic Evaluation}
We evaluate using standard ROUGE-1/2/L (\textbf{R-1/2/L}) and include a variation called \textbf{R1-c} that evaluates R1 on the reference summaries with entities not found in the input removed from the summary. 
We include the percentage of the base model summaries that are edited by the post-editor (\textbf{Edit\%}) and the following metrics:
\par 
\noindent \textbf{E-P$_{src}$} (\textbf{E-R$_{ref}$}) measures the percentage of entities in the generated summary (reference) present in the input (generated summary). 
E-P$_{src}$ performs on par with model-based, token-level metrics \cite{zhou-etal-2021-detecting,cao-etal-2022-hallucinated}. The subset of data without named entity errors has an E-P$_{src}$ of 100.
\par 
\noindent \textbf{BS-P$_{src}$ (BS-F1$_{ref}$)} represents the BERTScore \cite{zhang2019bertscore} precision (F1) w.r.t. the source article (reference summary).
\par 
\noindent \textbf{D$_{arc}$} measures the percentage of dependency arcs in summary entailed by the source article using the model from \citet{goyal-durrett-2021-annotating}.
\par 
\noindent \textbf{QAFE} is the QAFactEval question answering-based consistency metric  \citep{DBLP:journals/corr/abs-2112-08542}. 
\par 
\noindent \textbf{CoLA} To evaluate grammaticality, we apply a RoBERTa-large \cite{liu2019roberta} model from \citet{style20} trained on the CoLA dataset \cite{warstadt2018neural}, which includes sentences and labels for their grammatical acceptability.
\input{tables/human_eval_results}
\subsection{Results}
Results of applying post-editing models to BART on XSum are shown in Table \ref{table-main-result}, and examples, and results on CNN/DM, are in the Appendix.
\par
We note that BART-c improves across all factual consistency metrics compared to BART at the cost of ROUGE and other informativeness metrics. 
We find that SpanFact does not improve factual consistency, as this model is trained to fill in entity masks on the original, noisy dataset, but SpanFact-c improves factual consistency. 
While CCGS does improve factual consistency slightly, the model only edits a small percentage of the summaries (Edit\% of about 15). 
Other post-editors make a more realistic number of edits, considering that over 70\% of XSum reference summaries may contain factual inconsistencies \cite{maynez-etal-2020-faithfulness}.
\par
ReDRESS performs better than SpanFact-c on factual consistency and also E-R$_{ref}$, which aligns with its rewriting objective that can insert semantically relevant entities. 
FactPegasus improves entity precision by definition (it is not 100 due to differences in entity processing), but it is the only post-editor that decreases grammaticality by a noticeable margin.
Our CompEdit model improves factual consistency, and we see further improvements when applying our model on top of the ReDRESS corrector; ReDRESS as a first-stage corrector can insert suitable replacement entities while the second-stage compressor may remove any remaining entity errors.
CompEdit does not remove all entity errors in the summaries, and a qualitative inspection revealed that common world knowledge tokens, such as names of world leaders, were often left unchanged.  
Recent work has also noted the presence of extrinsic world knowledge errors \cite{cao-etal-2022-hallucinated}, and we leave a larger study of such artifacts to future work.
We note that vanilla ROUGE scores do decrease along with improvements in factual consistency. 
However, R1-c actually shows improvements when applying FactPegasus or CompEdit, showing that much of the loss in vanilla R1 is due to factual inconsistencies in the references.
% ReDRESS-based models show a decrease in R1-c, although the decrease is not as drastic compared to the vanilla R1 score. 
\par 
We show the results of applying the above post-editors on non-BART models in Table \ref{table-fasum_result_main}.
We find large improvements over \citet{zhu-etal-2021-enhancing}.
We see a gain in R1-c and entity precision when applying CompEdit to the pretrained UniLM, showing the benefits of our compression-based approach when the underlying model contains relatively high-quality summaries.
ReDRESS shows a large performance increase on BottomUp; its ability to completely rewrite the summary allows it to improve the non-pretrained, lower-quality summaries that require editing beyond just compression. 
There is also a benefit from the combination of ReDRESS and CompEdit in terms of entity precision for the BottomUp model, which has been shown to contain a high proportion of entity errors and inconsistencies \cite{huang-etal-2020-achieved,pagnoni-etal-2021-understanding}.
\par
Finally, we show the manually-annotated results of post-editing a non-pretrained model from the CNN/DM-based FactCC \cite{kryscinskiFactCC2019} test dataset in Table \ref{table-human_result}.
We compare with previously-reported results: an autoregressive post-editor trained on entity swaps and backtranslated paraphrases \cite{cao-etal-2020-factual}, and a method that substitutes entities from retrieved source sentences \cite{https://doi.org/10.48550/arxiv.2204.08263}.
As extrinsic entity errors, for which CompEdit and FactPegasus are designed, constitute only a small proportion of errors in this dataset, we only include ReDRESS + CompEdit.
CompEdit was applied to the ReDRESS outputs containing extrinsic named entity errors, 2.5\% of edited summaries, all of which were then labeled factually consistent.
% all of which remained consistent.
% 
% 
We see a large improvement in consistent summaries and similar trends in the benefit of ReDRESS and adding CompEdit as the above results on the non-pretrained BottomUp model.
% 
% We also note room for improvement in future work.

%% file: tables/fasum_results_main.tex
\begin{table}[t!]
\small
\centering
 \resizebox{\columnwidth}{!}{\begin{tabular}{|l | c c c | c c|}
\hline
Model & E-P$_{src}$ & BS-P$_{src}$ & D$_{arc}$ &   R1 & R1-c \\ \hline
UniLM & 60.63 & 42.45 & 0.83 & \textbf{42.13} & 40.74   \\ 
\hspace{1mm} FASumFC &  60.66 & 42.40 & 0.83 &  \textbf{42.17}  & 40.78  \\ 
\hspace{1mm} ReDRESS & 76.01 & \textbf{46.28} & 0.90 &  40.73 & 40.33   \\ 
\hspace{1mm} FactPegasus &  \textbf{99.06} & 43.28 & \textbf{0.91} &   39.26   & \textbf{41.05}  \\ 
\hspace{1mm} CompEdit (ours) &  \textbf{80.72} & 43.86 & 0.89 &  40.79  & \textbf{40.93}  \\ 
\hspace{1mm} ReDRESS + CompEdit & 78.44 & \textbf{45.43} & \textbf{0.92}  &  39.17  &   39.46  \\ \hline
% FASum & 71.01 & 37.17 & 0.71 &  30.27   &  29.90 \\ 
% \hspace{1mm} FASumFC & 68.59 & 37.03 & 0.71 &  30.19 & 29.83  \\ 
% \hspace{1mm} ReDRESS &  83.26 &  \textbf{44.56} & \textbf{0.88} & \textbf{32.76} & \textbf{32.91}  \\ 
% \hspace{1mm} FactPegasus &  \textbf{97.81} & 37.03 & 0.77 &  29.52  &  30.01  \\ 
% \hspace{1mm} CompEdit (ours) & \textbf{89.32} & 37.99 & 0.77 &  30.54  & 30.63  \\ 
% \hspace{1mm} ReDRESS + CompEdit &   83.04 & \textbf{43.17} &  \textbf{0.89}  &  \textbf{32.49} &  \textbf{32.88} \\ \hline
BottomUp & 43.31 & 31.38 & 0.41 & 26.90  &  26.28  \\ 
\hspace{1mm} FASumFC & 44.76 & 32.09 & 0.43 &  28.19  &  27.50  \\ 
\hspace{1mm} ReDRESS &   82.11 & \textbf{45.70} & \textbf{0.85} & \textbf{31.50} & \textbf{31.62}  \\ 
\hspace{1mm} FactPegasus &  \textbf{91.35} & 31.65 &  0.49 &  26.78  &  26.71 \\ 
\hspace{1mm} CompEdit (ours) & 79.74 & 34.13 &  0.54 & 27.65 &  28.40 \\ 
\hspace{1mm} ReDRESS + CompEdit &  \textbf{83.66}  & \textbf{44.30} & \textbf{0.87} &  \textbf{31.67} &  \textbf{32.03} \\ \hline
% TConvs2s & 70.58 & 36.93 & 0.72 & 31.45 &  31.00 \\ 
% \hspace{1mm} FASumFC & 69.53 & 36.57 & 0.73 & 32.43 &  31.95 \\ 
% \hspace{1mm} ReDRESS &  83.49 & \textbf{45.17} & \textbf{0.88} & \textbf{33.59} &  \textbf{33.71} \\ 
% \hspace{1mm} FactPegasus &  \textbf{97.56} & 37.44 & 0.77 &  30.57  & 31.08   \\ 
% \hspace{1mm} CompEdit (ours) & \textbf{88.08} & 38.80 & 0.79 & 32.04 &  32.06  \\ 
% \hspace{1mm} ReDRESS + CompEdit &  83.02  & \textbf{43.62} & \textbf{0.90} &  \textbf{33.17} &  \textbf{33.55} \\ \hline
\end{tabular}}
% \caption{Comparison of post-editors with FASumFC \cite{zhu-etal-2021-enhancing} on UniLM \cite{unilm} and FASum \cite{zhu-etal-2021-enhancing} outputs on XSum.} \label{table-fasum_result_part}
% \caption{Comparison of best-performing post-editors with FASumFC \cite{zhu-etal-2021-enhancing} on UniLM \cite{unilm}, FASum \cite{zhu-etal-2021-enhancing}, BottomUp \cite{gehrmann-etal-2018-bottom}, TConvS2S \cite{pmlr-v70-gehring17a} outputs on XSum.} \label{table-fasum_result_full}
\caption{Comparison of post-editors with FASumFC \cite{zhu-etal-2021-enhancing} on UniLM \cite{unilm} and BottomUp \cite{gehrmann-etal-2018-bottom} outputs on XSum.} \label{table-fasum_result_main}
\end{table}

%% file: tables/human_eval_results.tex
\begin{table}[t!]
\small
\centering
 \begin{tabular}{|l| c c |}
\hline
Model  & Consistent & Inconsistent \\ \hline
Unedited summaries & 441 & 62  \\ \hline
\citet{cao-etal-2020-factual} &  447 & 56  \\ 
\citet{https://doi.org/10.48550/arxiv.2204.08263} & 446 & 57 \\ 
ReDRESS + CompEdit & \textbf{473} & \textbf{30} \\ \hline
\end{tabular}
% 503 examples total
% 62 inconsistent - 47 (summaries edited) + 15 (remained inconsistent) = 30 (previously used 61).
\caption{Human evaluation of the number of factually consistent and inconsistent summaries after post-editing on the FactCC \cite{kryscinskiFactCC2019} test dataset.
} \label{table-human_result}
\end{table}

%% file: 4-conclusion.tex
\section{Conclusion}\label{sec:conclusion}
In this work, we propose a sentence-compression-based post-editing model to improve factual consistency in summarization while maintaining the informativeness and grammaticality of the resulting summaries. 
We show that this model can be used in tandem with a post-editor that performs extensive rewriting for further improvement, especially in pretrained models and datasets with a high proportion of entity errors. 
In future work, we plan to build on these models by studying the role of dataset artifacts in error correction and addressing unfixable summaries. 

%% file: 5-ethical-considerations.tex
\section{Limitations}\label{sec:limitations}
We train and test our models on publicly available news summarization datasets. 
Political and gender biases may exist in these datasets, and thus models trained on these datasets may propagate these biases. 
Furthermore, while our models are not language-specific, we only study English summarization. 
\par 
When used as intended, these post-editing models can help eliminate some factual inconsistencies in model summaries. 
However, these models do not remove all factual inconsistencies, and thus much care should be taken if one wants to use such models in a user-facing setting. 
\par
The experiments in this paper make use of A100 GPUs. 
We used up to 8 GPUs per experiment, and the experiments may take up to a day to run. 
Multiple experiments were run for each model, and future work should experiment with distilled models for more lightweight training. 
% 

%% file: 6-appendix.tex
\section{Additional Results}
Results of applying post-editing models to BART on CNN/DM are shown in Table \ref{table-main-result-app}.
BART-c decreases performance in factual consistency on CNN/DM, perhaps due to the reduction in training data points; CNN/DM is largely extractive and encourages factually consistent as-is.
This is reaffirmed by the small gap between SpanFact and SpanFact-c and the generally very high scores on CNN/DM, which leave less room for improvements compared to the XSum dataset. 
We also see a lower Edit\% on CNNDM, which aligns with \citet{pagnoni-etal-2021-understanding} where 27\% of BART summmaries on CNN/DM contained a factual inconsistency.
\par
In Table \ref{tab:examples}, we provide model outputs that illustrate the characteristics of the post-editors studied.
\par
We show the results of applying post-editors on additional non-BART models in Table \ref{table-fasum_result_app}.
We find similar trends as the post-editors applied to other non-pretrained in the main text.
\par
Additionally, we trained a post-editor only on sentence compression data; the model produces the compressed sentence (summary) given an uncompressed sentence (summary), with entities to remove marked with special tokens. 
This model applied to BART XSum summaries provided similar entity precision but resulted in about a 1.2 drop in ROUGE-1 and a five-point drop in entity recall on the validation set, so we did not include this model.
\section{Additional Model Details}
In this section, we provide additional details for our model, baselines, and metrics. 
To encourage retaining essential information, we filter data points from \citet{filippova-altun-2013-overcoming} in which the compressed sentences contain less than 75\% the number of tokens in the longer sentence.
We compared the effect of sentence compression data on downstream post-editor performance, experimenting with higher compression ratios as well as other sentence compression datasets \cite{clarke2008global}, but these models resulted either in lower ROUGE performance or a decrease in entity precision on the validation set.
To create training examples for our post-editor, we inserted one, two, and three entities into the references by applying our perturber. 
The size of the subsets of reference summaries entity precision of 100 on (training, validation) is (52k, 2.9k) for XSum and (160k, 6.5k) for CNN/DM.
We then sample 200k data points to train on a dataset of size similar to XSum and found that doubling the size of this data did not give further improvements. 
% 
% Other
\input{tables/results_app}

\input{tables/examples_analysis}

\input{tables/fasum_results_app}
We experimented with other approaches for marking extrinsic errors such as model-based approaches \citet{zhou-etal-2021-detecting} and \citet{deng-etal-2021-compression}, but the entity overlap approach performed better and is more interpretable. 
\par
We use the following inference parameters: (beam size, min generation length, max generation length, length penalty) for XSum = (6, 11, 62, 1.0) and CNN/DM = (4, 40, 140, 2.0). 
We retrain BART-large on XSum, and for CNN/DM we run inference from the fairseq/bart-large-cnn~\footnote{\url{https://huggingface.co/facebook/bart-large-cnn}}
checkpoint from the transformers library \cite{wolf-etal-2020-transformers}.
\par
For ReDRESS, we train BART-large rather than the BART-base model used in the original paper for consistency with the above models. 
A key component of this post-editor is the over-generation and reranking of the edited summaries; we rerank according to entity precision over the input.
We also vary how entities are matched in the ReDRESS retrieval component and use exact string matching to match how we filtered the clean data subset. 

For CCGS, we apply the XSum-trained reranker to both XSum and CNN/DM.
\par
For DAE, we apply the DAE\_xsum\_human\_best model.~\footnote{\url{https://drive.google.com/file/d/1TGS0RmS1sYlFyU52LxWfPA3Jiuo0d--M/}}
We provide our BERTScore run hash.~\footnote{bert-base-uncased\_L8\_no-idf\_version=0.3.9(hug\_trans=4.11.3)-rescaled}

%% file: tables/results_app.tex
\begin{table*}[!ht]
\small
\centering
\resizebox{\linewidth}{!}{\begin{tabular}{| l | c c  c  c | c c c c c c | c | c |}
\hline
Model & E-P$_{src}$ & BS-P$_{src}$ & D$_{arc}$ & QAFE  & E-R$_{ref}$ & BS-F1$_{ref}$ & R1 & R2 & RL & R1-c & CoLA & Edit\% \\ \hline
% \multicolumn{13}{|c|}{CNN/DM} \\
% \hline
BART & 98.50 & 71.10 & 98.57 & 4.550 & 58.93 & 43.55 & 44.05 & 21.12 & 41.02 & 44.11 & 95.68 & -  \\
BART-c & 97.60 & 68.59 & 97.88 & 4.405 & 58.00 & 43.70 & 44.09 & 20.89 & 41.11 & 44.14 & 95.11 & -  \\ \hline
SpanFact \tiny \cite{dong-etal-2020-multi} & 98.46 & 71.06 & 98.49 & 4.545 & 58.32 & 43.45 & 43.97 & 21.02 & 40.95 & 44.03 & \textbf{95.60} & 28.29 \\
SpanFact-c & 98.83 & 71.07 & 98.53 & 4.544 & 58.19 & 43.41 & 43.95 & 20.99 & 40.93 & 44.01 & 95.50 & 32.36 \\
CCGS \tiny \cite{chen-etal-2021-improving} & 98.18 & 71.04 & 98.49 & 4.466  & 58.00 & 43.48 & \textbf{43.98} & 21.03 & \textbf{40.96} & 44.04 & 95.48 & 14.93 \\
ReDRESS \tiny \cite{adams2022redress} & 99.03 & \textbf{71.32} & 98.66 & \textbf{4.557}  & \textbf{58.81} & \textbf{43.53} & \textbf{43.98} & \textbf{21.08} & \textbf{40.96} & 44.05 & \textbf{95.66} & \textbf{30.13} \\
FactPegasus \tiny \cite{wan2022factpegasus} & \textbf{99.94} & 71.10 & 98.68 & 4.543  & 58.40 & 43.44 & \textbf{43.98} & 21.03 & \textbf{40.96} & \textbf{44.12} & 95.21 & 13.44 \\
CompEdit (ours) & \textbf{99.10} & 71.13 & \textbf{98.75} & 4.554 & \textbf{58.61} & \textbf{43.50} & \textbf{44.02} & \textbf{21.08} & \textbf{41.01} & \textbf{44.11} & 95.52 & 27.93 \\
ReDRESS + CompEdit & 98.95 & \textbf{71.24} & \textbf{98.74} & \textbf{4.555} & 58.49 & 43.44 & 43.96 & 21.04 & 40.95 & 44.04 & 95.42  & \textbf{44.53} \\ \hline
\end{tabular}}
\caption{Baseline and post-editing automatic results for factual consistency, relevance, and grammaticality metrics on CNN/DM. The top two scores from the post-editors in each column are highlighted. 
% \jw{if needs more space we can put CNN results to appendix}
} \label{table-main-result-app}
\end{table*}

%% file: tables/examples_analysis.tex
\begin{table*}[!ht]
\small
\centering
 \resizebox{\textwidth}{!}{\begin{tabular}{p{0.35\linewidth}|p{0.1\linewidth}|p{0.55\linewidth}}
\hline
Source Article & System & Summary \\
\hline 
\multirow{3}{\linewidth}{Kevin Patrick Dawes, 33, was abducted in 2012 as he entered Syria. ... The Czech government, which represents US diplomatic interests in Syria, helped secure his release.} & BART & A US contractor who was held hostage in Syria \textcolor{red}{for more than three years} has been freed, the US State Department says. \\ \cline{2-3}
& ReDRESS, (w/CompEdit) & A US contractor who was held hostage in Syria has been freed \textcolor{green}{by Czech authorities}, the US State Department says. \\ \cline{2-3}
& FactPegasus, CompEdit & A US contractor who was held hostage in Syria has been freed, the US State Department says. \\ \hline 
\multirow{3}{\linewidth}{The victims were farming in an area... close proximity to Mount Sinabung. The volcano was still spewing ash on Sunday. ... Rescue teams are still scouring the area, looking for more victims who may have been killed or badly burned by the hot gas and ash clouds released in the eruption.} & BART & Rescue workers in Indonesia are searching for \textcolor{red}{more than 20 people feared dead} after the Mount Sinabung volcano erupted \textcolor{red}{on Saturday}.  \\ \cline{2-3}
& ReDRESS & Rescue workers in Indonesia are searching for more victims who may have been killed when the Mount Sinabung volcano erupted \textcolor{red}{on Saturday}. \\ \cline{2-3}
& ReDRESS + CompEdit & Rescue workers in Indonesia are searching for more victims who may have been killed when the Mount Sinabung volcano erupted. \\ \hline
\end{tabular}}
\caption{Example BART and post-editor outputs showing the capacity of ReDRESS to insert related entities and for CompEdit to remove errors. A comma separating two systems indicates that the two return the same summary.}
\label{tab:examples}
\end{table*}

%% file: tables/fasum_results_app.tex
\begin{table}[!ht]
\small
\centering
 \resizebox{\columnwidth}{!}{\begin{tabular}{|l | c c c | c c|}
\hline
% Model & E-P$_{src}$ & BS-P$_{src}$ & D$_{arc}$ &   R1 & R1-c \\ \hline
% UniLM & 60.63 & 42.45 & 0.83 & \textbf{42.13} & 40.74   \\ 
% \hspace{1mm} FASumFC &  60.66 & 42.40 & 0.83 &  \textbf{42.17}  & 40.78  \\ 
% \hspace{1mm} ReDRESS & 76.01 & \textbf{46.28} & 0.90 &  40.73 & 40.33   \\ 
% \hspace{1mm} FactPegasus &  \textbf{99.06} & 43.28 & \textbf{0.91} &   39.26   & \textbf{41.05}  \\ 
% \hspace{1mm} CompEdit (ours) &  \textbf{80.72} & 43.86 & 0.89 &  40.79  & \textbf{40.93}  \\ 
% \hspace{1mm} ReDRESS + CompEdit & 78.44 & \textbf{45.43} & \textbf{0.92}  &  39.17  &   39.46  \\ \hline
FASum & 71.01 & 37.17 & 0.71 &  30.27   &  29.90 \\ 
\hspace{1mm} FASumFC & 68.59 & 37.03 & 0.71 &  30.19 & 29.83  \\ 
\hspace{1mm} ReDRESS &  83.26 &  \textbf{44.56} & \textbf{0.88} & \textbf{32.76} & \textbf{32.91}  \\ 
\hspace{1mm} FactPegasus &  \textbf{97.81} & 37.03 & 0.77 &  29.52  &  30.01  \\ 
\hspace{1mm} CompEdit (ours) & \textbf{89.32} & 37.99 & 0.77 &  30.54  & 30.63  \\ 
\hspace{1mm} ReDRESS + CompEdit &   83.04 & \textbf{43.17} &  \textbf{0.89}  &  \textbf{32.49} &  \textbf{32.88} \\ \hline
% BottomUp & 43.31 & 31.38 & 0.41 & 26.90  &  26.28  \\ 
% \hspace{1mm} FASumFC & 44.76 & 32.09 & 0.43 &  28.19  &  27.50  \\ 
% \hspace{1mm} ReDRESS &   82.11 & \textbf{45.70} & \textbf{0.85} & \textbf{31.50} & \textbf{31.62}  \\ 
% \hspace{1mm} FactPegasus &  \textbf{91.35} & 31.65 &  0.49 &  26.78  &  26.71 \\ 
% \hspace{1mm} CompEdit (ours) & 79.74 & 34.13 &  0.54 & 27.65 &  28.40 \\ 
% \hspace{1mm} ReDRESS + CompEdit &  \textbf{83.66}  & \textbf{44.30} & \textbf{0.87} &  \textbf{31.67} &  \textbf{32.03} \\ \hline
TConvs2s & 70.58 & 36.93 & 0.72 & 31.45 &  31.00 \\ 
\hspace{1mm} FASumFC & 69.53 & 36.57 & 0.73 & 32.43 &  31.95 \\ 
\hspace{1mm} ReDRESS &  83.49 & \textbf{45.17} & \textbf{0.88} & \textbf{33.59} &  \textbf{33.71} \\ 
\hspace{1mm} FactPegasus &  \textbf{97.56} & 37.44 & 0.77 &  30.57  & 31.08   \\ 
\hspace{1mm} CompEdit (ours) & \textbf{88.08} & 38.80 & 0.79 & 32.04 &  32.06  \\ 
\hspace{1mm} ReDRESS + CompEdit &  83.02  & \textbf{43.62} & \textbf{0.90} &  \textbf{33.17} &  \textbf{33.55} \\ \hline
\end{tabular}}
% \caption{Comparison of best-performing post-editors with FASumFC \cite{zhu-etal-2021-enhancing} on UniLM \cite{unilm}, FASum \cite{zhu-etal-2021-enhancing}, BottomUp \cite{gehrmann-etal-2018-bottom}, TConvS2S \cite{pmlr-v70-gehring17a} outputs on XSum.} \label{table-fasum_result_full}
% \caption{Comparison of best-performing post-editors with FASumFC \cite{zhu-etal-2021-enhancing} on BottomUp \cite{gehrmann-etal-2018-bottom} and TConvS2S \cite{pmlr-v70-gehring17a} outputs on XSum.}
\caption{Comparison of post-editors with FASumFC  \cite{zhu-etal-2021-enhancing} on FASum \cite{zhu-etal-2021-enhancing} and TConvS2S \cite{pmlr-v70-gehring17a} outputs on XSum.} \label{table-fasum_result_app}
\end{table}